%% file: main.tex
\documentclass[sigconf]{acmart}

\usepackage{algorithm}
\usepackage{algpseudocode}

\algnewcommand\algorithmicforeach{\textbf{for each}}
\algdef{S}[FOR]{ForEach}[1]{\algorithmicforeach\ #1\ \algorithmicdo}

\usepackage{algpseudocode}
\AtBeginDocument{%
  \providecommand\BibTeX{{%
    \normalfont B\kern-0.5em{\scshape i\kern-0.25em b}\kern-0.8em\TeX}}}

\copyrightyear{2024}
\acmYear{2024}
\setcopyright{rightsretained}
\acmConference[GECCO '24]{Genetic and Evolutionary Computation Conference}{July 14--18, 2024}{Melbourne, VIC, Australia}
\acmBooktitle{Genetic and Evolutionary Computation Conference (GECCO '24), July 14--18, 2024, Melbourne, VIC, Australia}
\acmDOI{10.1145/3638529.3654116}
\acmISBN{979-8-4007-0494-9/24/07}

\acmSubmissionID{pap498}

\begin{document}
\author{Corinna Triebold}
\email{c.e.h.triebold@student.vu.nl}
\orcid{1234-5678-9012}\affiliation{%
  \institution{Vrije Universiteit Amsterdam}
  \city{Amsterdam}
  \country{The Netherlands}
}

\author{Anil Yaman}
\email{a.yaman@vu.nl}
\affiliation{%
  \institution{Vrije Universiteit Amsterdam}
  \city{Amsterdam}
  \country{The Netherlands}
}

\title{Evolving generalist controllers to handle a wide range of morphological variations}

\begin{abstract}

Neuro-evolutionary methods have proven effective in addressing a wide range of tasks. However, the study of the robustness and generalizability of evolved artificial neural networks (ANNs) has remained limited. This has immense implications in the fields like robotics where such controllers are used in control tasks. Unexpected morphological or environmental changes during operation can risk failure if the ANN controllers are unable to handle these changes. 
This paper proposes an algorithm that aims to enhance the robustness and generalizability of the controllers. This is achieved by introducing morphological variations during the evolutionary training process. As a results, it is possible to discover generalist controllers that can handle a wide range of morphological variations sufficiently without the need of the information regarding their morphologies or adaptation of their parameters. 
We perform an extensive experimental analysis on simulation that demonstrates the trade-off between specialist and generalist controllers. The results show that generalists are able to control a range of morphological variations with a cost of underperforming on a specific morphology relative to a specialist. This research contributes to the field by addressing the limited understanding of robustness and generalizability and proposes a method by which to improve these properties.

\end{abstract}



\begin{teaserfigure}
\centering
  \includegraphics[width=0.65\textwidth]{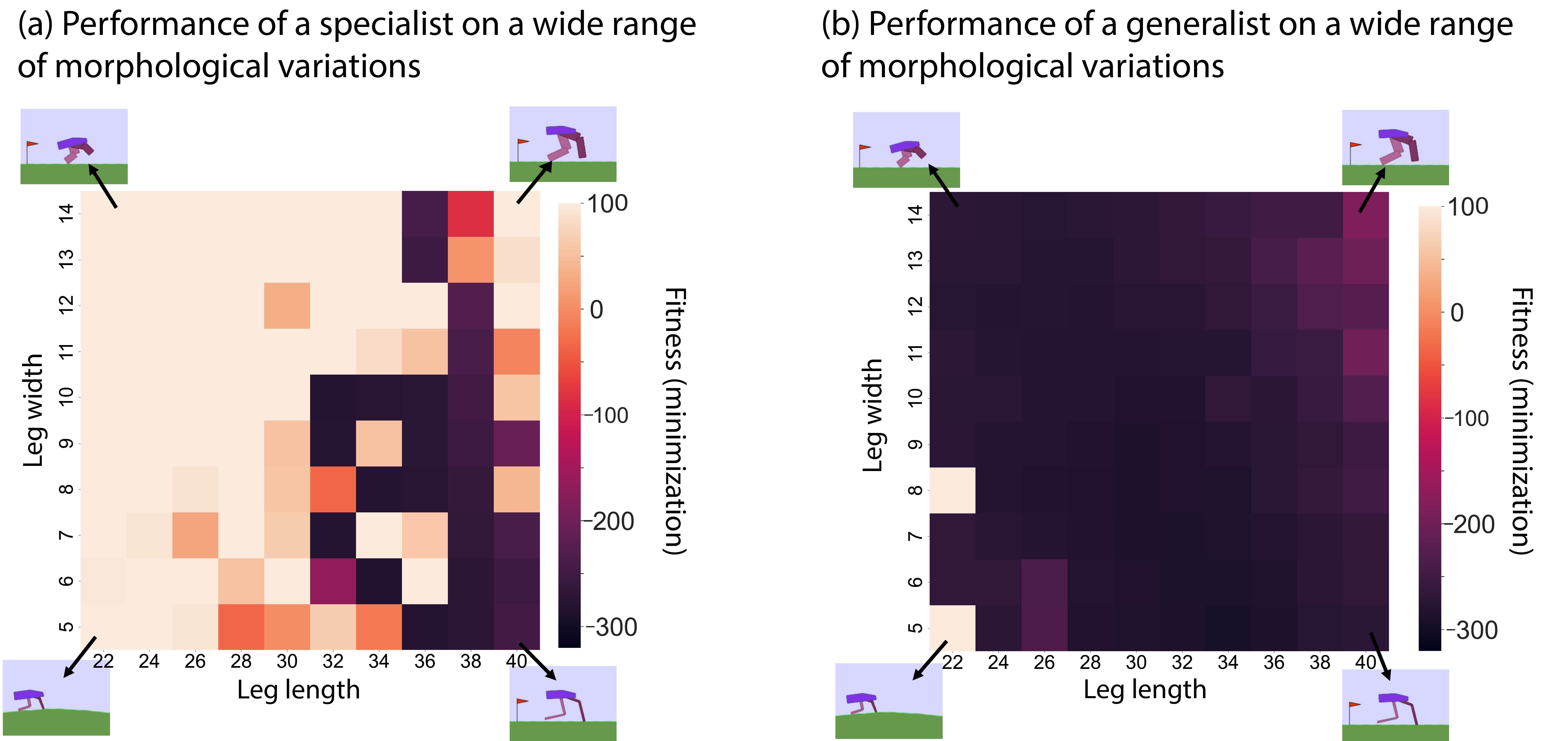}
  \caption{Controllers that are evolved on a single morphology (\textit{specialists}) show poor performance on a wider range of morphological variations (a), whereas, controllers that are introduced to a range or morphological variations during evolution (\textit{generalists}) show a higher performance on a wider range of morphological variations (b). In figures, each cell represents a morphological variation in the Bipedal Walker environment, parameterized by the leg length and width on the $x$ and $y$ axes.}
  \label{fig:teaser}
\end{teaserfigure}

\maketitle

\input{sections/intro}

\input{sections/background}
\input{sections/methods}

\input{sections/experimentalsetup}
\input{sections/results}
\input{sections/discussion}

\input{sections/conclusion}

\bibliographystyle{ACM-Reference-Format}
\bibliography{main.bbl}

\end{document}

%% file: sections/intro.tex
\section{Introduction}\label{s:intro}

In recent years, research has demonstrated the successful application of Evolution Strategies (ES) in optimizing Artificial Neural Networks (ANNs) for continuous control problems \cite{pagliuca2020efficacy, such2017deep}. These problems involve complex dynamics and require controllers adept at handling continuous variables and actions (e.g. in OpenAI gym simulation environments \cite{brockman2016openai,salimans2017evolution}). However, promoting generalizability and robustness of these ANNs remains an important ongoing research effort especially for ensuring stable and reliable performance in conditions that are not encountered during the evolutionary optimization process~\cite{valentim2022adversarial, mangal, carvalho2023role, rajeswaran2017towards}. 

Typically, ANN controllers are optimized for a specific agent morphology in a specific environment \cite{pagliuca2020efficacy}. Here, the agent's morphology refers to its physical structure, for example, the legs, joints, and hull configuration of the OpenAI gym Bipedal Walker. This can lead to specialization, causing the controller to ``overfit'' to the morphology and lower its performance for morphological variations such as changes to the leg length and width. Specialization is usually desired since it can promote better performance, however, it may not be desirable in cases where there is a possibility of encountering conditions different to the ones encountered during evolution. This can, for instance, be due to damage~\citep{putter2017evolving, cully2015robots, risi2013confronting}. In these situations, retraining the agent may be too costly or impossible~\cite{putter2017evolving, 8628627}. Thus, a generalist, a morphologically robust ANN, that can handle changes without significant performance sacrifices~\cite{putter2017evolving} can be desirable. An example illustration of performances of specialized vs. generalized controllers on morphological variations of Bipedal Walker is given in Figure~\ref{fig:teaser}. In (a), we can observe poor generalizability of a specialist controller that is evolved on single morphology, whereas, in (b), we observe the performance of a generalist, evolved using the algorithm proposed in this work, that shows better performance on a wide range of variations.

Current approaches to handling morphological or environmental change  rely on learning behavioral repertoires~\citep{cully2015robots} or online adaptation~\citep{yaman2021evolving, nygaard2021real, risi2013confronting, 8628627, putter2017evolving}. However, if the adaptation relies on self-diagnosis based on sensor data, sensor malfunction can lead to potential issues~\citep{nygaard2021real, risi2013confronting, 8628627, putter2017evolving}. Retraining a new controller is an alternative but consumes time and computational power, and in real-world applications, can risk the operation. A limited number of works have explored robustness of evolved controllers to morphological variations and mostly focused on sensory-motor variations \citep{carvalho2023role, 8628627}. Larger-scale investigations into the robustness concerning physical morphological variations are necessary.

To address this need, this paper introduces an approach that aims to promote the evolution of generalist controllers. We hypothesize that introducing morphological variations during the evolutionary process can lead to the evolution of generalist controllers. This draws on principles from domain randomization and curriculum learning (CL) \cite{narvekar2020curriculum, akkaya2019solving}. CL has been applied in the context of neuro-evolution, but the focus has been on task difficulty or co-evolving a morphology \cite{milano2021automated, wang2022curriculum, wang2019poet}. However, our aim is to achieve robustness across variations without increasing the difficulty of presented morphologies. We further expect that generalizability comes with a tradeoff in specialization on a particular morphology ~\citep{geman1992neural}. A specialist is likely to have high variance resulting in a poor generalization (i.e. \textit{overfitting}), whereas, a generalist is likely to have high bias resulting in worse performance on a single morphology (i.e. \textit{underfitting}). Striking the right balance between specialization vs. generalization is crucial.  To test this hypothesis, we analysed the relation between the amount of variation introduced during the evolutionary processes versus the level of robustness and generalization achieved. Furthermore, we introduce evolutionary branching, akin to speciation in nature, to partition the morphology space into smaller clusters (''species'') to balance the tradeoff between performance and generalizability. Additionally, we hypothesize, that similar to CL, the order that this variation is introduced in also makes an impact. Thus, we propose and test three ``training schedules'', namely, incremental, random and random walk (with two parameters).

We perform an extensive analysis on our algorithm on four test cases in the OpenAI gym environment. Our results confirm that incrementally introducing morphological variations promotes the emergence of generalist controllers. We observe that generalist controllers exhibit better mean performance on a wide range of morphological variations relative to the specialists. They are more robust and generalize to unseen variations. However, on a particular morphology that a specialist evolved for the generalists perform comparatively worse. Thus, the generalist controllers can be highly advantageous in situations where unexpected situations are likely to arise, they can maintain their performance within their limits of generalization without the need of adaptation. From previous literature, we know that morphological differences can have significant impact on the behavior, it may not be possible to find a controller that can produce a generalizable behavior for all~\citep{pfeifer2006body}. Despite this, our focus is on identifying a generalist controller that is effective across various morphologies with a sufficient degree.  Overall, our approach is highly supported by the theoretical foundations of learning \cite{raviv2022variability,geman1992neural,yang2020rethinking}, and contributes to the further understanding of the tradeoff between specialization vs. generalization.

%% file: sections/background.tex
\section{Background}\label{s:background}
This section provides a background on reletad concepts and literature involving neuro-evolution for continuous control problems, robustness, and generalizability in ANN controllers, and a discussion on evolutionary ensembles and branching.

\subsection{Neuro-evolution for continuous control problems}

Neuro-evolution is effective for optimizing ANNs, especially when dealing with non-differentiable, sparse, and objective functions that are not easily defined~\citep{stanley2019designing, floreano2008neuroevolution, Miikkulainen2010}. In continuous control problems, the use of large ANN controllers is common due to the complexity of handling action and state variables in a continuous domain. However, scaling up the controller size increases the optimization challenge, especially in black-box approaches without gradient information. Natural Evolution Strategies (NES) address this by estimating gradient information, making them advantageous for optimization in complex high-dimensional problems~\citep{wierstra2014natural}.

NES involves sampling candidate solutions from the sampling distribution, evaluating their fitness via an objective function, and using these results to estimate the gradient (natural gradient). The sampling distribution is then updated based on this information and the process is repeated till a stopping condition is met ~\citep{glasmachers2010exponential, wierstra2014natural}. Among the existing NES algorithms, xNES demonstrated to be one of the most powerful variants~\citep{schaul2012benchmarking, glasmachers2010exponential}.

\subsection{Robustness and Generalizability}

ANN robustness refers to its ability to maintain performance despite variations or perturbations in input data. This can be categorized into local and global robustness. Local robustness ensures consistent output within a certain radius ($\delta$) for a given input ($x$) and its perturbations ($x^{\prime}$). Global robustness extends this property to all inputs~\citep{mangal}. In control tasks and robotics, robustness is crucial, covering aspects like environmental, sensory, and adaptive robustness~\citep{miki2022learning}. Morphological robustness involves sustaining task performance despite variations in agent morphology~\citep{putter2017evolving}. This is different from morphological adaptiveness, which focuses on morphology adaptation~\citep{nygaard2021real}. Generalizability refers to an ANN's ability to transfer performance to unencountered situations~\citep{rajeswaran2017towards}. Morphological generalizability ensures task performance across unseen morphologies. These concepts are interconnected in that robust ANNs can be expected to generalize better. Achieving these qualities is crucial in robotics, enabling agents to handle uncertainties, adapt to changes, and maintain consistent task performance in diverse conditions. 

\subsection{Evolutionary ensembles}
Ensemble learning methods make use of a divide and conquer approach. Many problems are simply too complex to be solved through the use of a single ANN \citep{liu2000evolutionary}. By combining multiple networks the ensemble is able to break the problem into simpler sub-problems which are easier to solve. This process typically has three steps: member generation, selection, and  combination. Ensemble learning methods have been found to be more robust and better at generalizing~\citep{dong2020survey}. 
Evolutionary ensembles take the three steps and treat them as optimization problems to which evolutionary computing can be applied~\citep{telikani2021evolutionary}. Evolutionary computing methods are well-suited for generating ensemble members due to the fact that they generate pools of candidate solutions by design. The inherent selection mechanisms facilitate easy member selection. To ensure the effectiveness of evolutionary ensemble generation, it is crucial to pursue diversity among ensemble members by using different training data-sets, different learning algorithms, etc. These approaches can enhance the ensemble's overall performance and robustness \citep{telikani2021evolutionary}.

\subsection{Evolutionary branching}
Evolutionary branching is a natural phenomenon where a previously homogeneous population splits into two or more distinct clusters \citep{geritz1997dynamics,doebeli2000evolutionary}. Directional selection occurs when the environment of the population favors one phenotype allowing the individuals that exhibit this trait to survive and reproduce more often. Frequency-dependent selection can occur in both negative and positive  directions. Positive frequency selection occurs when a phenotype or genotype increases in fitness because it is more common within the population. In contrast, negative frequency selection describes the process of increasing fitness when the presence of a phenotype or genotype in the population becomes less common \cite{brisson2018negative}. Branching can also come about through isolation mechanisms that divide a population and prevent any further exchange of genetic material. Over time these selection and isolation mechanisms ultimately result in the formation of multiple distinct lineages or clusters within the population. Evolutionary branching leads to better adaption for the organisms and increased biodiversity \cite{charlesworth1997effects}. 

%% file: sections/methods.tex
\section{Evolving generalist controllers}\label{sec:methods}

This section outlines the algorithm we propose to promote robustness and generalization of evolved ANN controllers. The pseudocode is provided in Algorithm \ref{algo}. The code is accessible publicly\footnote{ \url{https://github.com/cotrieo/evolving-generalist-controllers}}. Although, it is possible to apply our algorithm for other forms of variation (e.g. environmental), here, we consider morphological variations. These variations are introduced during the evolutionary process, enabling the discovery of generalist controllers capable of effectively controlling a range of morphological variations without the need of being informed by the morphology parameters.

Consider a set of training morphologies $M = \{m_1, m_2, \dots, m_n\}$, and a set of generalist solutions $G=\{I_1, I_2, \dots I_k\}$ where $k\leq n$. Formally, the process of finding generalist controllers can be defined as, finding $G$ containing a minimum number of controllers where each can perform the control task sufficiently well for one or more morphologies in $M$. We assume no overlap in morphology assignments. The more morphologies a controller $I_i$ controls sufficiently well, the more generalist it is. Ideally, when $k=1$, there is only a single generalist controller that achieves control of all morphological variations. The performance of the controllers $I_i$ on the control tasks of a morphology set $K\subseteq M$ is defined as the average performance of the control task performance of $I_i$ evaluated on all morphologies in $K$, denoted as: 
\begin{equation}
    \hat{f} = F(I_i,K) = \frac{1}{|K|}\sum^{|K|}_{j=1} \text{EVAL}(I_i, m_j)  
    \label{eq:fitness}
\end{equation}
where $m_j\in K$ and EVAL is the evaluation function of controller $I_i$ on $m_j$.

We employ xNES in our experiments, but the algorithm is adaptable to any standard evolutionary algorithm. The hyperparameters include a predefined fixed learning set $M$ of  morphologies and a training schedule $T$ that introduces a morphology in $M$ in a particular order in each generation during the evolutionary process. The algorithm initializes with an empty set of $G$ and an empty set $O$ (Lines 1-2), to include outlier morphologies during the evolutionary process. $O$ is used for evolutionary branching.

The initial population is generated using uniform random initialization from a domain range, and individuals evaluated on morphology $m_j$ provided by the schedule $T$. For each generation (including the first), the best individual $I_{best}$ in $P$ is selected based on the fitness value on $m_j$, and evaluated on all morphologies in set $M$ and its average fitness $\hat{f}^{\prime}$ is found (Line 14). In case of the first generation, $I_{best}$ is saved as $I_i$, its fitness $\hat{f}^{\prime}$ saved as $\hat{f}$, for the following generations, we check if $\hat{f}^{\prime} < \hat{f_1}$ (assuming minimization problem), then, $I_i$ and $\hat{f}$ are replaced by $I_{best}$ and $\hat{f_1}^{\prime}$ (Lines 15-17).

If the fitness does not improve for $h$ number of generations (Line 20), then morphologies where the current $I_1$ is not able to perform well are removed from $M$ and added to $O$. This is determined by identifying the morphologies where fitness is lower than one standard deviation (denoted as $\sigma$ in Lines 23 and 25) from the mean fitness on all morphologies (Lines 21-27). This leads to \textit{evolutionary branching} where, later, a new controller is evolved for $O$ since the current generalist controller on $M$ cannot achieve a good performance. Other thresholds (i.e. two or three standard deviations) were tested but this reduced the probability of removing the outliers so good overall performance could not be maintained. Smaller thresholds on the other hand, lead to more specialists since the number of morphologies considered to be outliers will increase.

\begin{algorithm}
\caption{Evolving generalist controllers}\label{euclid}
\begin{algorithmic}[1]
\State $M$ = set of $n$ morphologies 
\State $G$ = $\{\}$
\While{$n \ != 0$}
    \State $O$ = $\{\}$
    \State $\hat{f}$ = inf \Comment{Assuming minimization}
    \State $j=0$ \Comment{Counter for the morphology $m_j$} 
    \State $n$ = $|M|$ 
    
    \While{$\hat{f} >$ max fitness and evals $<$ max evaluations}
        \State generate new solution candidates
        \State evaluate candidates on $m_j$
        \State $I_{best}$ = individual with highest fitness
        \State $f_{best}$ = fitness of $I_{best}$
        \State update search distribution parameters \Comment{Assuming ES}
        \State $\hat{f}^{\prime}$ = mean fitness of $I_{best}$ on $M$ \Comment{According to Eq.\eqref{eq:fitness}}
        \If{$\hat{f}^{\prime} < \hat{f}$} 
        
            \State $I_i$ = $I_{best}$
            \State $\hat{f} =  \hat{f}^{\prime}$
        \EndIf
        \State $j$ = $j+1$ mod $n$ \Comment{Increment morphology counter}
        \If{$f_{best}$ has not improved for $h$ generations} 
        
            \ForEach{$m$ in $M$}
                \State fitness = evaluate $I_i$ on $m$
                \If{$fitness  <= (\hat{f}_i + \sigma)$}
                    \State $n^{\prime} =  n^{\prime} + 1$
                \Else{ $fitness > (\hat{f}_i + \sigma)$}
                    \State add $m$ to $O$
                    \State remove $m$ from $M$
                \EndIf
            \EndFor
        \EndIf
        \If{$n^{\prime} = n$}
            \State break
        \Else
            \State $n = n^{\prime}$
        \EndIf
    \EndWhile
\State $M = O$
\State add $I_i$ to $G$
\EndWhile
\State return $G$
\end{algorithmic}
\label{algo}
\end{algorithm}

Following removal, the process continues on $M$ in the same way until a satisfactory average fitness is found or the maximum number of evaluations is reached (Line 8). At the end of this process, $M$ is replaced by $O$ and $I_i$ is added to $G$ (Lines 37-38). The whole process is repeated (but this time for the outlier set $O$) until no more morphologies are left. Restarting the evolutionary process eliminates any information from the previous evolution. A clean slate allows the evolution to optimize without the bias of the previous generations. The whole process is repeated until maximum number of evaluations is reached or there are no morphologies left in $O$. 

Overall, this process of removing the morphologies that cannot be generalized, partitions the learning space. Different ANN controller ``species'' evolve for various regions of the morphology space if a single ANN controller cannot perform well on the whole space. Thus, the algorithm produces a set of ANN controllers that are evolved to control certain regions of the morphology space well, and taken together forms an ANN controller ensemble for the whole morphology space. Ideally, the goal would be to find a single controller that can handle the variations on the whole morphological space considered, however, then the overall performance may suffer since there is always tradeoff between specialization vs. generalization.

%% file: sections/experimentalsetup.tex
\section{Experimental setup}\label{s:evaluation}

\subsection{Morphological learning sets}
We evaluate our algorithm on four control problems within the OpenAI gym environment~\citep{brockman2016openai}. For each problem, we define 1, 16, 25, 49, and 64 morphological learning sets ($M$) to examine the impact of learning set size on generating robust and generalizable controllers. All learning sets start from the same initial parameters and use identical increment steps, resulting in varying final parameters based on set size. Despite using small incremental steps, these changes significantly affect the agents' morphologies. Each problem-learning set size combination was run 30 times for statistical significance.

\begin{enumerate}
\item CartPole: Balancing a pole attached to a cart. Morphological variation parameters were the pole length and mass, starting from 0.1 and increasing by 0.1 increments.

\item Bipedal Walker: 2D bipedal locomotion task. Variation parameters include leg length and width, starting from 5 and 22 with increments of 2 and 1, respectively.

\item Ant: 3D Mujoco locomotion task of a quadruped agent. Variation parameters include lengths of upper and lower legs, starting from 0.2 and 0.4 in increments of 0.1.

\item Walker2D: 2D Mujoco locomotion of a bipedal agent. Variation parameters include length of upper leg, starting from 0.225 in increments of 0.025, and lower leg starting at 0.25 in increments of 0.05.
    
\end{enumerate}

\subsection{Experimental parameters}
All ANNs share the same fully connected single-layer feedforward topology and real-valued vectors represent the ANN parameters during evolution. Smaller single-layer networks have proven effective for similar continuous control problems, with no significant performance loss \citep{pagliuca2020efficacy}. The use of multilayered networks did not demonstrate added benefits, and a simpler topology with fewer parameters accelerates runtime \citep{rajeswaran2017towards}. The Tanh activation function is applied to all neurons for compatibility with motor torques in the range [-1, 1], allowing direct use of ANN outputs as input for the agent's next step. Parameters for each problem: 
\begin{itemize}
    \item CartPole: 121 connection weights (4 input, 20 hidden, 1 output neuron, bias of 1). 
 \item Bipedal Walker: 584 connection weights (24 input, 20 hidden, 4 output neurons, bias of 1).
\item Ant: 728 connection weights (27 input, 20 hidden, 8 output neurons, bias of 1).
\item Walker2d: 246 connection weights (17 input, 10 hidden, 6 output neurons,  bias of 1).
\end{itemize}
 
The experiments employed the xNES algorithm, running for a maximum of 5000 generations with a population size determined by the default settings: $N = 4 + \lfloor(3 * \log(l) \rfloor$, where $l$ represents the search space dimensions~\citep{evotorch2023arxiv}. The maximum number of generations also includes the restarts for the outlier sets ($O$) in case of evolutionary branching. 
The desired fitness was determined by the problem rewards, resulting in different reward ranges. For Bipedal Walker 280 points were considered as good performance (it gets close to its goal coordinates). CartPole, Ant, and Walker2D have no set maximum points; evolution halted upon reaching predefined fitness thresholds to manage run times. Thresholds were established through testing: 800 for CartPole (pole is upright 80\% of the time), 1800 for Ant (end-of-plane traversal), and 1800 for Walker2D (significant forward movement). Ant and Walker2D's default reward functions included a health reward, set to 0 here to prevent exploitation by standing still. All problems were treated as minimization problems, with results presented accordingly.


\subsection{Training schedules}
We conducted additional experiments on CartPole and Bipedal Walker to explore training schedules introducing morphological variations. Three schedules were tested: incremental, random, and random walk, with two parameter settings for random walk, totaling 4 experiments. In the incremental schedule, variations are introduced sequentially in a 2D parameter space, starting from initial $x$ and $y$ morphological parameters, incrementing along the $x$ axis first, then the $y$ axis. The random schedule samples $x$ and $y$ parameters with equal probabilities. The random walk starts from random $x$ and $y$ values, selecting the next variation from the neighborhood with equal probability, with step sizes of 1 and 5.

\subsection{Evaluation metrics}\label{sec:metrics}
To assess ANN controller robustness and generalizability, we evaluate average fitness performance on three sets: \textbf{Default morphology} is the performance on an arbitrarily selected morphology based on default settings in the OpenAI environment. \textbf{Local variations set} contains morphologies that are in and around the neighborhood of the default morphology (robustness). This set introduces unseen variations within a distance of 6 from the learning morphologies, ensuring a minimum of 36 additional morphologies. The larger the learning set is, the larger its local test set. \textbf{Global variations set} contains morphological variations beyond the local test set. This test space is consistent for all ANNs, providing a broad measure of generalizability regardless of the learning set size.



%% file: sections/results.tex
\section{Results}
In this section, we will present the results for each problem, along with observations on training schedule experiments, branching, and fitness trends. Visual representations of generalist controller behavior are available online.\footnote{ \url{https://youtu.be/eew4X5gBvLQ?si=29G5iBIpB5xhJEaT}}.

\subsection{Problem Outcomes}
\subsubsection{CartPole}
\begin{figure} [ht!]
    \centering
    \includegraphics[ width=0.8\linewidth]{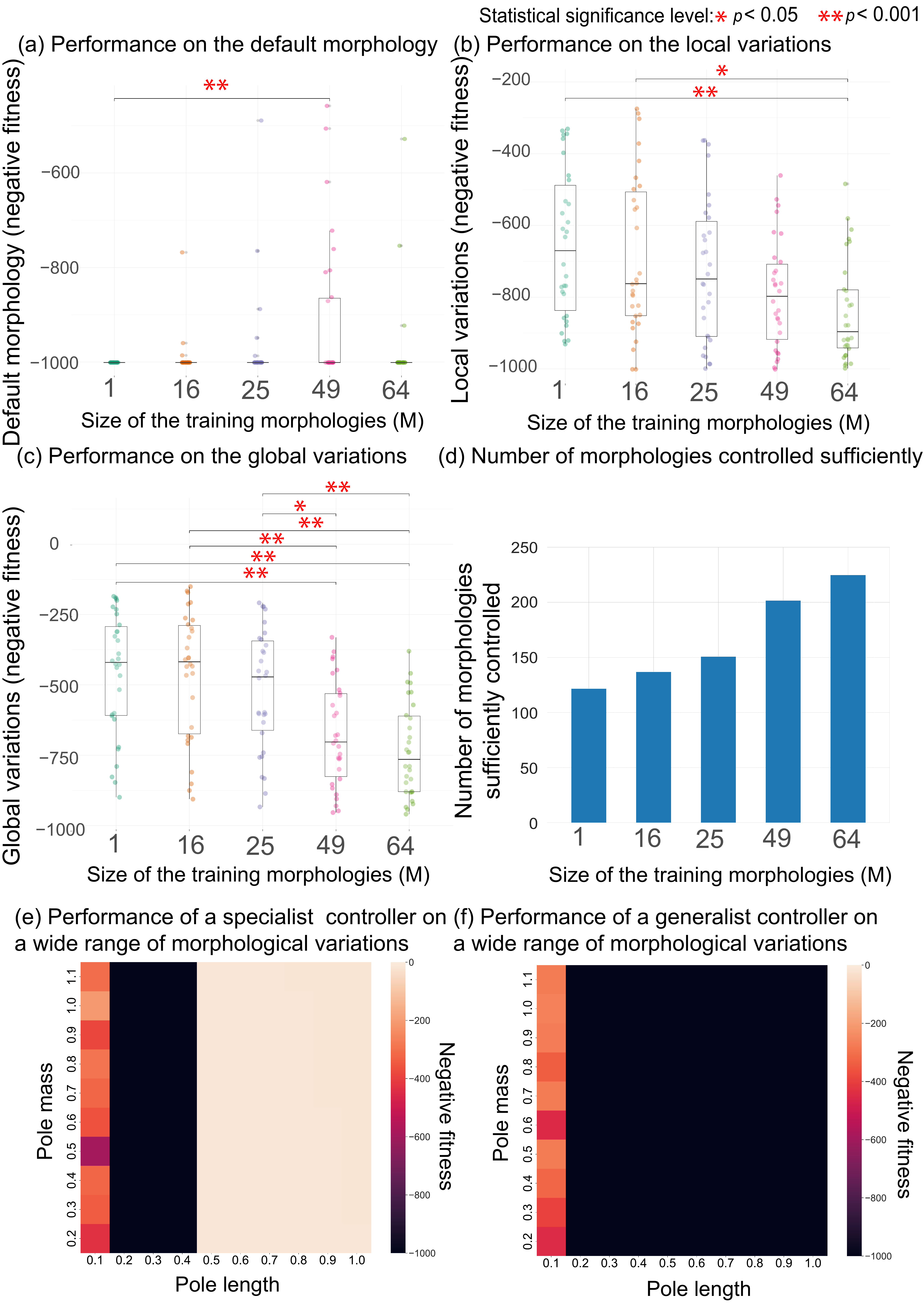}
    \caption{CartPole, specialists excel on the default morphology but perform poorly on local and global variations compared to generalists.}
    \label{fig:cartpoleResults}
\end{figure}

Due to the task's simplicity, no branching occurred, allowing a single generalist controller to effectively handle all learning set sizes. On the default morphology (Figure~\ref{fig:cartpoleResults} (a)), performance shows increased variance with larger learning set sizes, yet all achieve a median representing the best possible score. Conversely, for the local variations set (Figure~\ref{fig:cartpoleResults} (b)), larger learning sets lead to improved performance, indicating robustness to small perturbations. Significant differences exist between controllers evolved on different learning set sizes, with larger sets skewing towards better performance. On the global variations set (Figure~\ref{fig:cartpoleResults} (c)), controllers exhibit improved generalist performance. Statistically significant differences are evident between most learning set sizes. In Figure~\ref{fig:cartpoleResults} (d), as the learning set size grows, controllers handle an increasing number of morphologies sufficiently well, from 37.35\% to 69.14\% of the global set. Figures~\ref{fig:cartpoleResults} (e) and (f) depict performance on the global set for controllers evolved with learning set sizes of 1 and 64. The controller in (e), evolved on 1 morphology, struggles with over 50\% of the morphological variations, while the generalist demonstrates better performance across a broader range of variations.

\subsubsection{Bipedal Walker}

\begin{figure}
    \centering
    \includegraphics[ width=0.8\linewidth]{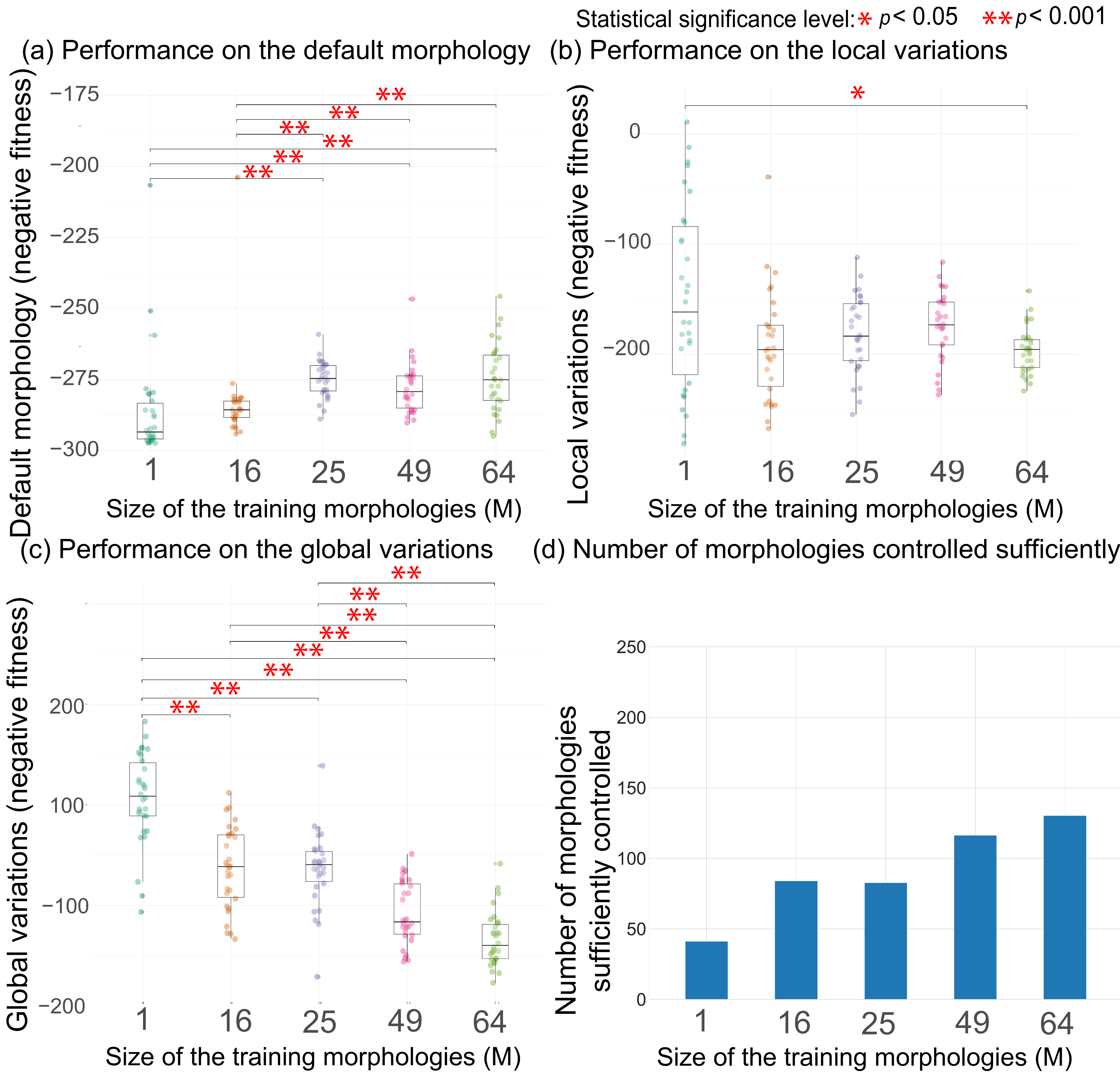}
    \caption{Bipedal Walker, specialists excel on the default morphology but perform poorly on local and global variations compared to generalists.}
    \label{fig:bipedalResults}
\end{figure}

For this problem, evolutionary branching was common, indicating that it's complexity requires multiple ANNs to manage larger morphological spaces effectively. In Figure~\ref{fig:bipedalResults} (a), performance on the default morphology decreases with increasing learning set size, suggesting a negative impact. Statistical analysis strongly supports this observation. This is intuitive since in the cases of evolution with more variations, the time spent on different variations could be spent on the default morphology, which would improve the performance for this morphology.

On the local variations set in Figure~\ref{fig:bipedalResults} (b), a reverse trend to (a) is observed, with median performances ranging from -161.64 to -195.55. As set size increases, performance only slightly improves, but variance significantly decreases indicating improved robustness. On the global variations set (Figure~\ref{fig:bipedalResults} (c)), the reverse trend to (a) is even more pronounced, with a strong downward trend indicating increased performance and decreased variance.
Both the local and global results show statistical significant differences between the learning set sizes. Figure~\ref{fig:bipedalResults} (d) illustrates the increasing number of morphologies effectively controlled by controllers evolved with larger learning sets.

\subsubsection{Ant}

\begin{figure}
    \centering
    \includegraphics[ width=0.8\linewidth]{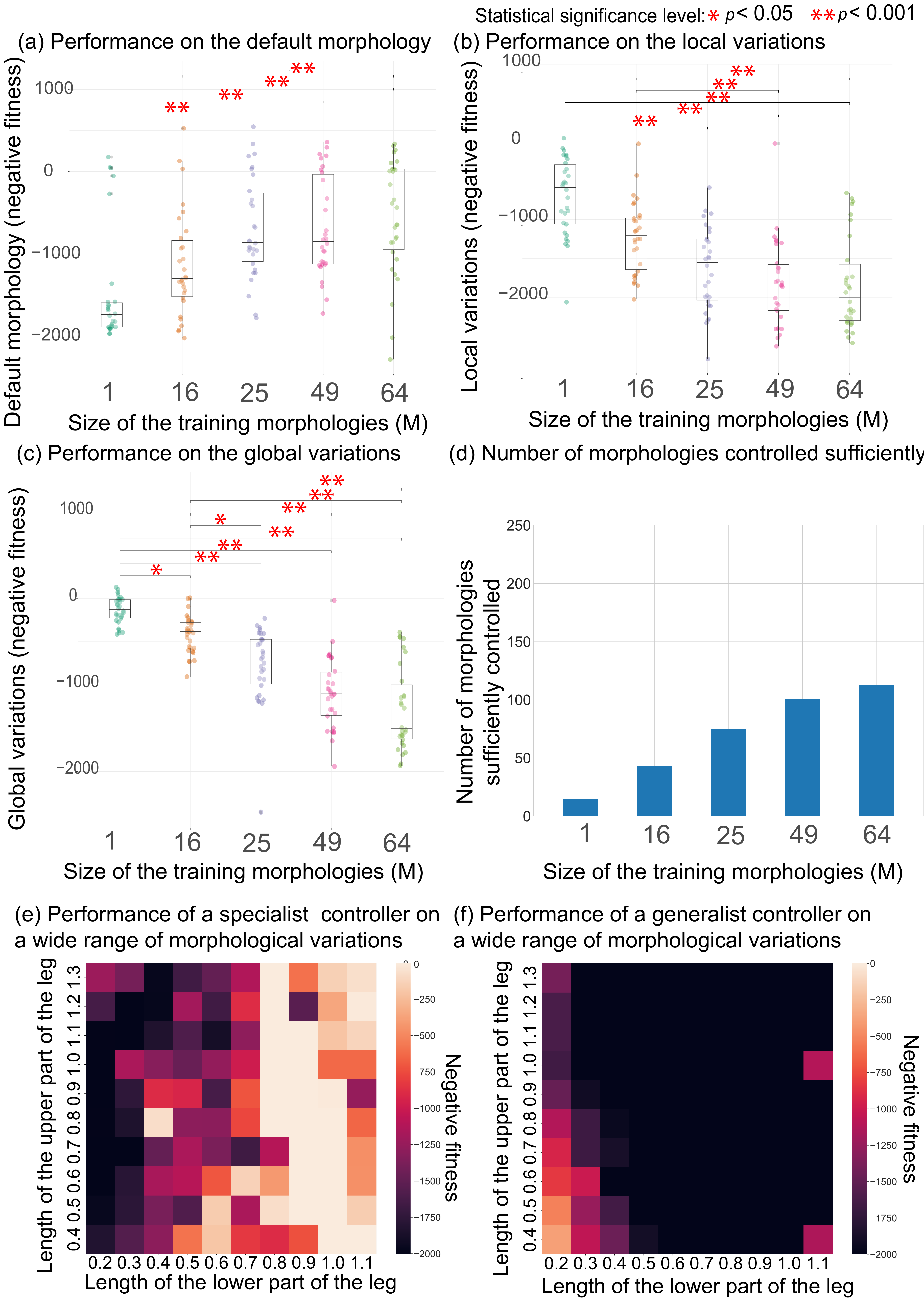}
    \caption{Ant, specialists excel on the default morphology but perform poorly on other variations.}
    \label{fig:antResults}
\end{figure}

Due to the Ant's problem complexity, runs with large learning sets exhibited evolutionary branching. In Figure~\ref{fig:antResults} (a), a clear upward trend in median fitness highlights a negative impact of larger learning sets on performance on the default morphology. Controllers evolved on the default achieved an median fitness of -1482.11, contrasting with -562.48 for those exposed to the most variations, showing statistically significant differences.

On the local variations set (Figure~\ref{fig:antResults} (b)), a steep downward trend indicates improved performance, opposite to the trend in (a). Median scores for controllers evolved on default and 64 morphologies were -587.78 and -1998.08 respectively, with a statistically significant difference. Similarly, Figure~\ref{fig:antResults} (c) illustrates a steep downward trend in the global variations set as the learning set size increases, also statistically significant. Figure~\ref{fig:antResults} (d) shows an increasing trend, growing from 14 to 112 morphologies. Figures~\ref{fig:antResults} (e) and (f) depict controllers evolved on the default and 64 morphologies respectively, highlighting the latter's greater generalizability in handling a wide range of morphological variations.

\begin{figure} 
    \centering
    \includegraphics[ width=0.8\linewidth]{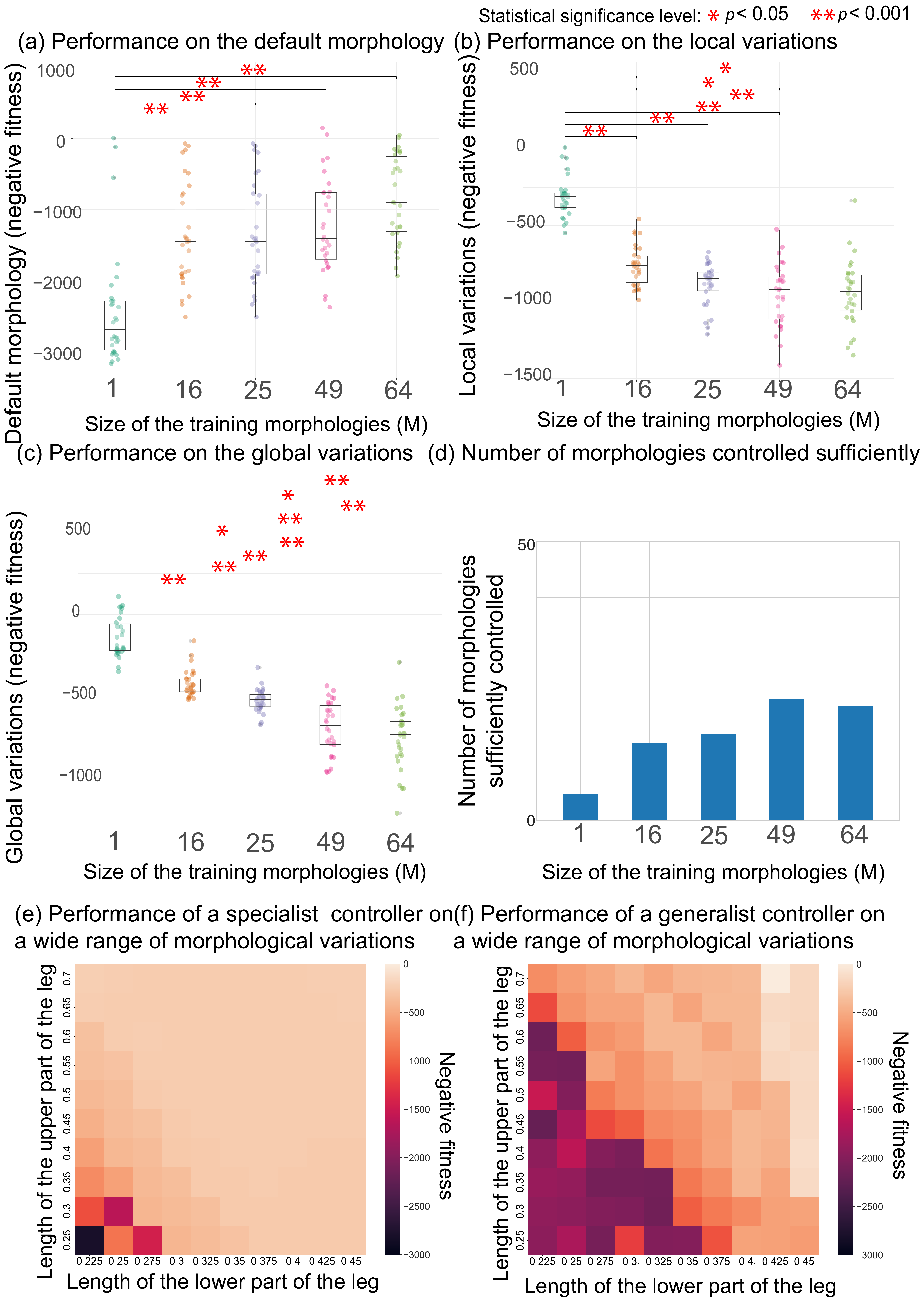}
    \caption{Walker2D, specialists excel on the default morphology but perform poorly on other variations.}
    \label{fig:walkerResults}
\end{figure}

\subsubsection{Walker2D}

In Figure~\ref{fig:walkerResults} (a), a clear upward trend on the default morphology, indicates deteriorating performance with increasing learning sets. The average fitness sharply jumps from -2420.34 to -843.47 between set sizes 1 and 64, showcasing the challenge of controlling the default Walker2D with a generalist controller. On the local variations set (Figure~\ref{fig:walkerResults} (b)), performance improves as the learning morphologies increase, contrasting with the default morphology trend in (a). Controllers on default and 64 morphologies achieve median performance between -311.59 and -930.12, with statistically significant $p$-values. A similar and stronger improvement trend is observed in the global set (Figure~\ref{fig:walkerResults} (c)).
Figure~\ref{fig:walkerResults} (d) demonstrates that as the learning set size increases, controllers handle more morphologies effectively, although this increase is less pronounced compared to other problems. In Figures~\ref{fig:walkerResults} (e) and (f), controllers evolved with a larger set demonstrate better performance across a range of variations compared to those evolved on the default morphology. Figure~\ref{fig:walkerResults} (f) illustrates the task's difficulty, as generalist controllers do not transfer to other variations as well as in other problems.

\subsection{Fitness Trends and Branching}

\begin{figure} 
    \centering
    \includegraphics[width=0.8\linewidth]{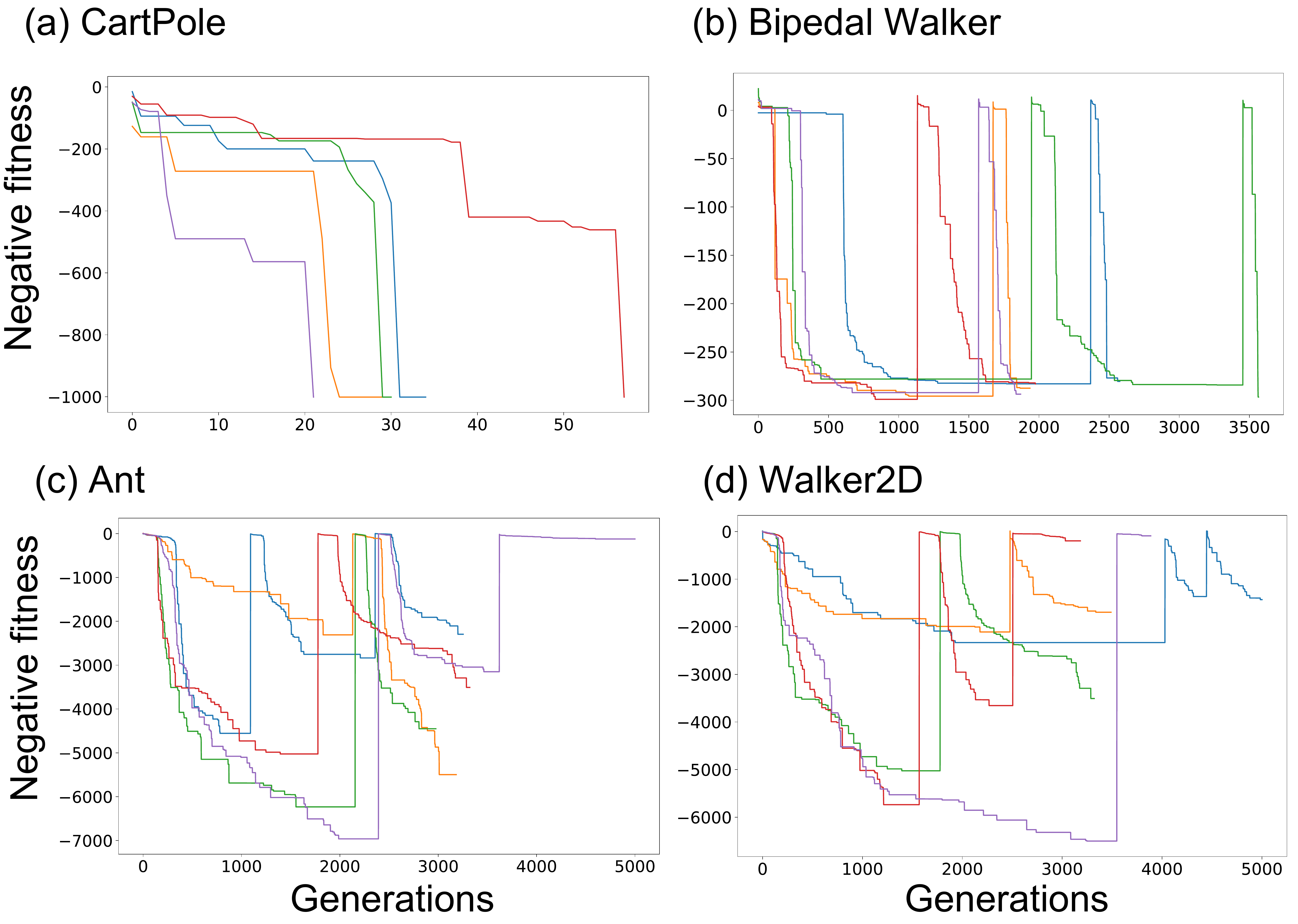}
    \caption{Illustration of fitness progression of five independent evolutionary processes (on morphology set size of 64).}
    \label{fig:evolutionaryRuns}
\end{figure}

Figure~\ref{fig:evolutionaryRuns} shows representative fitness trends during the evolutionary processes for each problem with the maximal training set size 64. For CartPole no evolutionary branching was observed: A single generalist controller is learned less than 100 generations. In the other problems, we can observe occasional fitness increases that indicate the restart of the evolutionary process following branching. Table~\ref{tab:clusters} shows the average number of evolutionary branches (and standard deviation) observed based on the size of training morphologies for each problem type. All problems, aside from CartPole, show an increase in the number of clusters as the training set grows supporting the notion that larger morphology sets need to be partitioned to achieve desired performance. Despite the increasing size of the training set and branching, the average number of generations required to find a solution did not increase at the same pace. Notably, between the specialist and the maximal generalist (with training set sizes of 1 and 64 respectively, resulting in a relative increase of 6300\%), the average number of generations only demonstrated a relative increase of 68.22\% for CartPole, 124.84\% for the Bipedal Walker, 92\% for the Ant, and 47.84\% for the Walker2D. This suggests that the algorithm offers speed benefits compared to evolving a specialist for each morphology separately.

\begin{table} [ht!]
\centering
\footnotesize
\caption{The table displays the mean number of clusters (and standard deviation) for all problems and learning set sizes.}\label{tab:clusters}
\begin{tabular}{|l|c|c|c|c|c|}
\hline
\begin{tabular}[c]{@{}l@{}}\textbf{}\end{tabular} & \textbf{1} & \textbf{16}  & \textbf{25}  & \textbf{49} & \textbf{64}                                                      \\ \hline
\textbf{CartPole}                                                                                      & \begin{tabular}[c]{@{}c@{}}1 \end{tabular} & \begin{tabular}[c]{@{}c@{}}1   \end{tabular}       & \begin{tabular}[c]{@{}c@{}}1   \end{tabular}        & \begin{tabular}[c]{@{}c@{}}1   \end{tabular}        & \begin{tabular}[c]{@{}c@{}}1   \end{tabular}       \\ \hline
\textbf{Bipedal W.}                                                                                & \begin{tabular}[c]{@{}c@{}}1   \end{tabular} & \begin{tabular}[c]{@{}c@{}}1.67   $\pm$ 0.96\end{tabular} & \begin{tabular}[c]{@{}c@{}}1.367   $\pm$ 0.56\end{tabular} & \begin{tabular}[c]{@{}c@{}}2.367   $\pm$ 0.61\end{tabular} & \begin{tabular}[c]{@{}c@{}}2.53   $\pm$ 0.78\end{tabular}  \\ \hline
\textbf{Ant}                                                                                          & \begin{tabular}[c]{@{}c@{}}1   \end{tabular} & \begin{tabular}[c]{@{}c@{}}1.67   $\pm$ 0.84\end{tabular} & \begin{tabular}[c]{@{}c@{}}1.7   $\pm$ 0.92\end{tabular}   & \begin{tabular}[c]{@{}c@{}}1.9  $\pm$ 1.12\end{tabular}    & \begin{tabular}[c]{@{}c@{}}2.33   $\pm$ 1.24\end{tabular} \\ \hline
\textbf{Walker2D}                                                                                      & \begin{tabular}[c]{@{}c@{}}1   \end{tabular} & \begin{tabular}[c]{@{}c@{}}2.4   $\pm$ 1.04\end{tabular}   & \begin{tabular}[c]{@{}c@{}}2.9   $\pm$ 0.99\end{tabular}   & \begin{tabular}[c]{@{}c@{}}3.2   $\pm$ 1.06\end{tabular}   & \begin{tabular}[c]{@{}c@{}}3.1   $\pm$ 1.24\end{tabular}  \\ \hline
\end{tabular}
\end{table}

\subsection{Baseline Comparison}

The proposed approach was compared to a baseline algorithm that aims to optimize a generalist controller for all morphology variations. We used standard xNES implementation and computed fitness in each generation as finding the average performance on all morphologies. Note that the proposed approach performs an evaluation on a single morphology in each generation but switches the morphology between generations. In addition, it finds average fitness of the best individual on all morphologies. Therefore, in each generation, it performs $k + n$ evaluations where $k$ and $n$ are population size and number of morphologies respectively. On the other hand, the baseline algorithm performs $k\times n$ evaluations in each generations. 

\begin{figure} 
    \centering
    \includegraphics[width=0.8\linewidth]{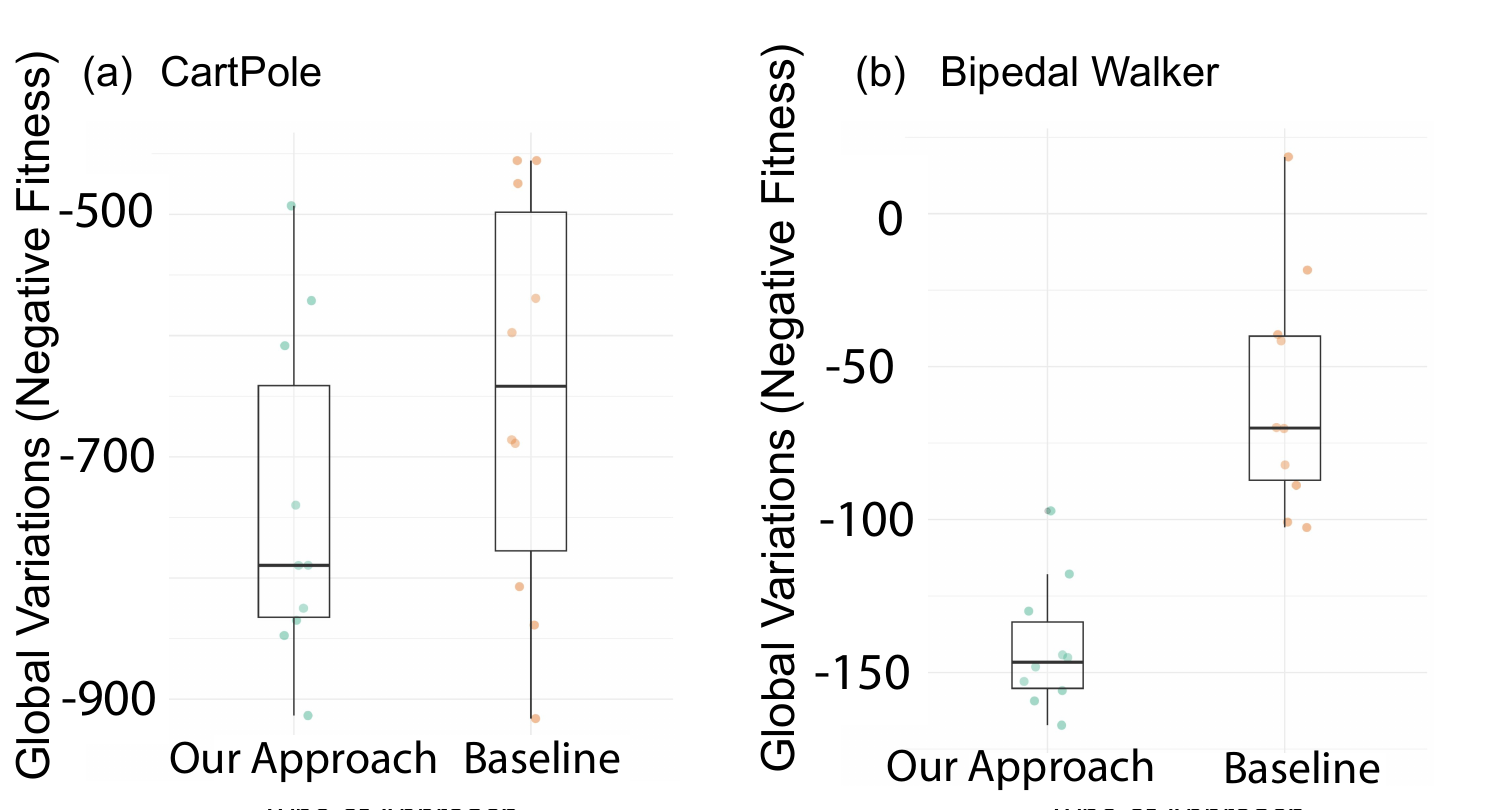}
    \caption{Proposed algorithm performs significantly better than the baseline algorithm on CartPole and Bipedal Walker.}
    \label{fig:addEXP}
\end{figure}

As depicted in Figure \ref{fig:addEXP}, the proposed approach demonstrated a superior performance compared to the baseline for CartPole (a) and Bipedal Walker (b) tasks. For instance, on Bipedal Walker, the proposed approach achieved the presented results around 1.7e+5 evaluations (around 1954 generations) whereas the baseline algorithm achieved the results at around 2.9e+5 evaluations (around 200 generations).

\subsection{Training schedules}

\begin{figure} 
    \centering
    \includegraphics[width=0.8\linewidth]{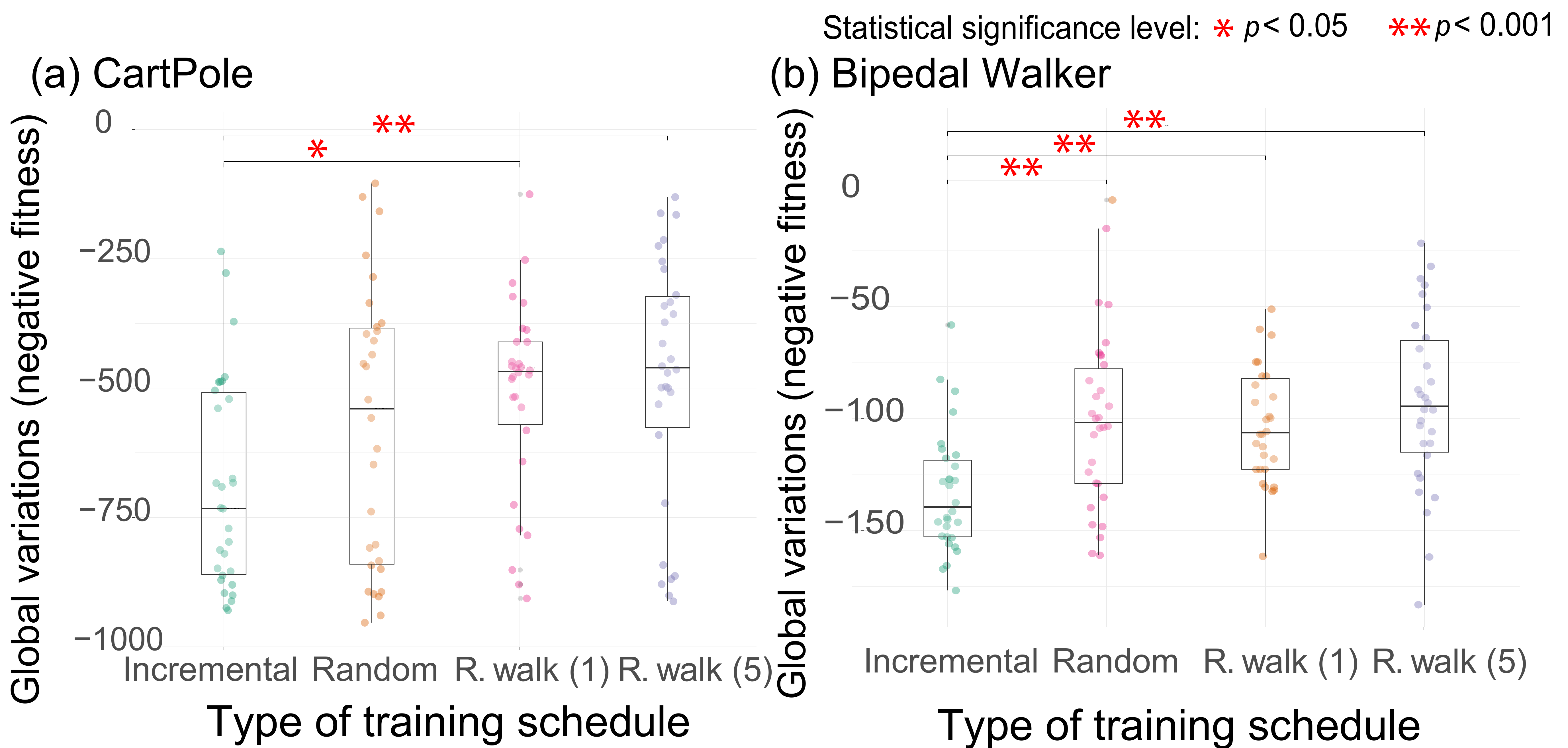}
    \caption{The incremental training schedule achieves the best results relative to other training schedules.}
    \label{fig:trainingSchedules}
\end{figure}

Figure~\ref{fig:trainingSchedules} depicts the results of four experiments using incremental, random, and the random walks on morphology size 64 for the CartPole and Bipedal Walker. The incremental training schedule demonstrates the best generalization performance, likely attributed to its ordered introduction of morphologies with only small morphological differences. In contrast, the random schedule introduces morphologies uniformly but with random order, potentially causing large jumps in morphological differences during evolution. Random walk with a step size of 1 may exhibit bias in morphology selection, but the differences are limited. However, in random walk with a step size of 5, biases may occur but in tandem with larger morphological differences resulting in the worst generalization performance.

%% file: sections/discussion.tex
\section{Discussion}\label{s:discussion}
The results strongly support the notion that controllers evolved across a broader set of morphological variations exhibit better robustness and generalization compared to those evolved on fewer variations. However, a tradeoff exists, controllers evolved on fewer variations excel in specific morphologies lose out when transferred to other morphologies, while those with larger learning sets perform well across a broader range but may underperform in specific morphologies, earning them labels of specialists or generalists  based on their morphological range. Although this observation holds across problems, differences exist, particularly in the Walker2D problem, suggesting problem-specific influences on generalist evolution. The trade-off is less pronounced in number of morphologies that are handled sufficiently well (which remained lower than 30 in all cases) but more evident in average fitness which shows drastic differences. Due to the inherent fragile design of this problem, it is more difficult to achieve control of the walker. This tradeoff aligns with expectations based on the bias-variance trade-off, revealing that increased generalization comes at the cost of specialization, the cost of this varies across problems. The order of morphological variations introduced during evolution plays a crucial role, with the incremental training schedule proving most effective. Experimentation with dynamic training schedules could offer further enhancements. As shown in Section 5.3, a baseline that aims to find a generalist controller by maximizing the average performance of a controller on all morphologies achieved drastically lower performance relative to the proposed approach. This is due to the fact that, our approach is way more efficient since it performs less number of evaluations per generation and introduce variation between generations.

While the our algorithm is capable of evolving robust and generalist ANN controllers, limitations exist. The current approach treats generalists as a by-product of specialist evolution. While this approach is successful, it may be possible to pursue the evolution of generalist controllers through multi-objective optimization, by treating each objective as performance on a morphological variation \citep{deb2002fast, gupta2015multifactorial}. However, this approach would also likely to face scalability issues similar to the ones observed in 5.3 due to the computational expense of evaluating each controller on all objectives. Our approach offers computational efficiency by limiting generalist evaluation to once per generation. To address domination issues, we utilize evolutionary branching, inspired by natural evolution, which divides the morphology space into subsets, evolving separate controllers when a single generalist is insufficient. This can lead to the emergence of specialized controllers and a partitioned morphology space, akin to an ANN ensemble with each member assigned to control a subset (or ''species'') of the morphological space. Evolutionary branching in nature occurs when different traits offer fitness advantages in different ecological niches \citep{doebeli2000evolutionary}.


Further investigations into the algorithm by utilizing other measures, like the median, to assess the generalizability during evolution could reveal interesting insights. Additionally, the algorithm currently does not employ other strategies to enhance generalization. Techniques like weight decay, as demonstrated by \citep{salimans2017evolution}, could be adapted. Alternatively, lexicase selection or a limited evaluation approach could be incorporated ~\citep{morse2016simple,yaman2018limited}, where generalist controllers are determined based on the ``inhereted'' fitness over the morphological variations rather than specifically evaluating them on the entire learning set. 

Other limitations of this work include the reliance on a pre-determined set of hand-generated morphologies. This can still introduce boundaries to the generalizability. Furthermore, the experiments only included variations that effect the limbs equally. Real world changes can result in unequal morphologies. Therefore, the experimental setup could be extended to cover such cases. The algorithm could also be tested on other forms of variations like environmental or task variation. This work also only considered a single ANN topology. Another interesting direction is to look into various types of ANNs and topologies, like LSTMs, to assess the impact of these.

In evolutionary robotics ANN controllers are evolved for real-world robotic tasks \citep{doncieux2015evolutionary}. Controllers are typically learned in a simulated environment and then transferred to the real world, this results in a challenge known as the reality gap \citep{koos2012transferability, doncieux2015evolutionary, jakobi1995noise}. Besides handling unexpected situations generalist controllers can also help reduce this gap. Therefore, progress toward robustness and generalizability is crucial for effectively addressing existing challenges.

%% file: sections/conclusion.tex
\section{Conclusion}\label{s:conclusion}

This paper examines the robustness and generalization of evolved ANN controllers in control tasks, with a focus on morphological variations. We introduce an algorithm that evolves ANN controllers capable of handling diverse morphologies without specific adjustments during operation. The evolved generalist behavior eliminates the need for morphology-specific learning or parameter adaptation during operation. This is achieved by introducing morphological variations during evolution.

Results show a tradeoff between the number of introduced morphological variations and performance. Controllers facing fewer variations specialize in specific morphologies but lack transferability. Conversely, those exposed to more variations exhibit superior performance across diverse and unseen morphologies. The training schedule significantly influences results, with the incremental schedule yielding networks with the highest robustness and generalization.

%% file: main.bbl

\begin{thebibliography}{45}


\ifx \showCODEN    \undefined \def \showCODEN     #1{\unskip}     \fi
\ifx \showDOI      \undefined \def \showDOI       #1{#1}\fi
\ifx \showISBNx    \undefined \def \showISBNx     #1{\unskip}     \fi
\ifx \showISBNxiii \undefined \def \showISBNxiii  #1{\unskip}     \fi
\ifx \showISSN     \undefined \def \showISSN      #1{\unskip}     \fi
\ifx \showLCCN     \undefined \def \showLCCN      #1{\unskip}     \fi
\ifx \shownote     \undefined \def \shownote      #1{#1}          \fi
\ifx \showarticletitle \undefined \def \showarticletitle #1{#1}   \fi
\ifx \showURL      \undefined \def \showURL       {\relax}        \fi
\providecommand\bibfield[2]{#2}
\providecommand\bibinfo[2]{#2}
\providecommand\natexlab[1]{#1}
\providecommand\showeprint[2][]{arXiv:#2}

\bibitem[\protect\citeauthoryear{Akkaya, Andrychowicz, Chociej, Litwin, McGrew, Petron, Paino, Plappert, Powell, Ribas, et~al\mbox{.}}{Akkaya et~al\mbox{.}}{2019}]%
        {akkaya2019solving}
\bibfield{author}{\bibinfo{person}{Ilge Akkaya}, \bibinfo{person}{Marcin Andrychowicz}, \bibinfo{person}{Maciek Chociej}, \bibinfo{person}{Mateusz Litwin}, \bibinfo{person}{Bob McGrew}, \bibinfo{person}{Arthur Petron}, \bibinfo{person}{Alex Paino}, \bibinfo{person}{Matthias Plappert}, \bibinfo{person}{Glenn Powell}, \bibinfo{person}{Raphael Ribas}, {et~al\mbox{.}}} \bibinfo{year}{2019}\natexlab{}.
\newblock \showarticletitle{Solving rubik's cube with a robot hand}.
\newblock \bibinfo{journal}{\emph{arXiv preprint arXiv:1910.07113}} (\bibinfo{year}{2019}).
\newblock


\bibitem[\protect\citeauthoryear{Brisson}{Brisson}{2018}]%
        {brisson2018negative}
\bibfield{author}{\bibinfo{person}{Dustin Brisson}.} \bibinfo{year}{2018}\natexlab{}.
\newblock \showarticletitle{Negative frequency-dependent selection is frequently confounding}.
\newblock \bibinfo{journal}{\emph{Frontiers in Ecology and Evolution}}  \bibinfo{volume}{6} (\bibinfo{year}{2018}), \bibinfo{pages}{10}.
\newblock


\bibitem[\protect\citeauthoryear{Brockman, Cheung, Pettersson, Schneider, Schulman, Tang, and Zaremba}{Brockman et~al\mbox{.}}{2016}]%
        {brockman2016openai}
\bibfield{author}{\bibinfo{person}{Greg Brockman}, \bibinfo{person}{Vicki Cheung}, \bibinfo{person}{Ludwig Pettersson}, \bibinfo{person}{Jonas Schneider}, \bibinfo{person}{John Schulman}, \bibinfo{person}{Jie Tang}, {and} \bibinfo{person}{Wojciech Zaremba}.} \bibinfo{year}{2016}\natexlab{}.
\newblock \showarticletitle{Openai gym}.
\newblock \bibinfo{journal}{\emph{arXiv preprint arXiv:1606.01540}} (\bibinfo{year}{2016}).
\newblock


\bibitem[\protect\citeauthoryear{Carvalho and Nolfi}{Carvalho and Nolfi}{2023}]%
        {carvalho2023role}
\bibfield{author}{\bibinfo{person}{Jonata~Tyska Carvalho} {and} \bibinfo{person}{Stefano Nolfi}.} \bibinfo{year}{2023}\natexlab{}.
\newblock \showarticletitle{The Role of Morphological Variation in Evolutionary Robotics: Maximizing Performance and Robustness}.
\newblock \bibinfo{journal}{\emph{Evolutionary Computation}} (\bibinfo{year}{2023}), \bibinfo{pages}{1--18}.
\newblock


\bibitem[\protect\citeauthoryear{Charlesworth, Nordborg, and Charlesworth}{Charlesworth et~al\mbox{.}}{1997}]%
        {charlesworth1997effects}
\bibfield{author}{\bibinfo{person}{Brian Charlesworth}, \bibinfo{person}{Magnus Nordborg}, {and} \bibinfo{person}{Deborah Charlesworth}.} \bibinfo{year}{1997}\natexlab{}.
\newblock \showarticletitle{The effects of local selection, balanced polymorphism and background selection on equilibrium patterns of genetic diversity in subdivided populations}.
\newblock \bibinfo{journal}{\emph{Genetics Research}} \bibinfo{volume}{70}, \bibinfo{number}{2} (\bibinfo{year}{1997}), \bibinfo{pages}{155--174}.
\newblock


\bibitem[\protect\citeauthoryear{Cully, Clune, Tarapore, and Mouret}{Cully et~al\mbox{.}}{2015}]%
        {cully2015robots}
\bibfield{author}{\bibinfo{person}{Antoine Cully}, \bibinfo{person}{Jeff Clune}, \bibinfo{person}{Danesh Tarapore}, {and} \bibinfo{person}{Jean-Baptiste Mouret}.} \bibinfo{year}{2015}\natexlab{}.
\newblock \showarticletitle{Robots that can adapt like animals}.
\newblock \bibinfo{journal}{\emph{Nature}} \bibinfo{volume}{521}, \bibinfo{number}{7553} (\bibinfo{year}{2015}), \bibinfo{pages}{503--507}.
\newblock


\bibitem[\protect\citeauthoryear{Deb, Pratap, Agarwal, and Meyarivan}{Deb et~al\mbox{.}}{2002}]%
        {deb2002fast}
\bibfield{author}{\bibinfo{person}{Kalyanmoy Deb}, \bibinfo{person}{Amrit Pratap}, \bibinfo{person}{Sameer Agarwal}, {and} \bibinfo{person}{TAMT Meyarivan}.} \bibinfo{year}{2002}\natexlab{}.
\newblock \showarticletitle{A fast and elitist multiobjective genetic algorithm: NSGA-II}.
\newblock \bibinfo{journal}{\emph{IEEE transactions on evolutionary computation}} \bibinfo{volume}{6}, \bibinfo{number}{2} (\bibinfo{year}{2002}), \bibinfo{pages}{182--197}.
\newblock


\bibitem[\protect\citeauthoryear{Doebeli and Dieckmann}{Doebeli and Dieckmann}{2000}]%
        {doebeli2000evolutionary}
\bibfield{author}{\bibinfo{person}{Michael Doebeli} {and} \bibinfo{person}{Ulf Dieckmann}.} \bibinfo{year}{2000}\natexlab{}.
\newblock \showarticletitle{Evolutionary branching and sympatric speciation caused by different types of ecological interactions}.
\newblock \bibinfo{journal}{\emph{The american naturalist}} \bibinfo{volume}{156}, \bibinfo{number}{S4} (\bibinfo{year}{2000}), \bibinfo{pages}{S77--S101}.
\newblock


\bibitem[\protect\citeauthoryear{Doncieux, Bredeche, Mouret, and Eiben}{Doncieux et~al\mbox{.}}{2015}]%
        {doncieux2015evolutionary}
\bibfield{author}{\bibinfo{person}{Stephane Doncieux}, \bibinfo{person}{Nicolas Bredeche}, \bibinfo{person}{Jean-Baptiste Mouret}, {and} \bibinfo{person}{Agoston~E Eiben}.} \bibinfo{year}{2015}\natexlab{}.
\newblock \showarticletitle{Evolutionary robotics: what, why, and where to}.
\newblock \bibinfo{journal}{\emph{Frontiers in Robotics and AI}}  \bibinfo{volume}{2} (\bibinfo{year}{2015}), \bibinfo{pages}{4}.
\newblock


\bibitem[\protect\citeauthoryear{Dong, Yu, Cao, Shi, and Ma}{Dong et~al\mbox{.}}{2020}]%
        {dong2020survey}
\bibfield{author}{\bibinfo{person}{Xibin Dong}, \bibinfo{person}{Zhiwen Yu}, \bibinfo{person}{Wenming Cao}, \bibinfo{person}{Yifan Shi}, {and} \bibinfo{person}{Qianli Ma}.} \bibinfo{year}{2020}\natexlab{}.
\newblock \showarticletitle{A survey on ensemble learning}.
\newblock \bibinfo{journal}{\emph{Frontiers of Computer Science}}  \bibinfo{volume}{14} (\bibinfo{year}{2020}), \bibinfo{pages}{241--258}.
\newblock


\bibitem[\protect\citeauthoryear{Floreano, D{\"u}rr, and Mattiussi}{Floreano et~al\mbox{.}}{2008}]%
        {floreano2008neuroevolution}
\bibfield{author}{\bibinfo{person}{Dario Floreano}, \bibinfo{person}{Peter D{\"u}rr}, {and} \bibinfo{person}{Claudio Mattiussi}.} \bibinfo{year}{2008}\natexlab{}.
\newblock \showarticletitle{Neuroevolution: from architectures to learning}.
\newblock \bibinfo{journal}{\emph{Evolutionary intelligence}}  \bibinfo{volume}{1} (\bibinfo{year}{2008}), \bibinfo{pages}{47--62}.
\newblock


\bibitem[\protect\citeauthoryear{Geman, Bienenstock, and Doursat}{Geman et~al\mbox{.}}{1992}]%
        {geman1992neural}
\bibfield{author}{\bibinfo{person}{Stuart Geman}, \bibinfo{person}{Elie Bienenstock}, {and} \bibinfo{person}{Ren{\'e} Doursat}.} \bibinfo{year}{1992}\natexlab{}.
\newblock \showarticletitle{Neural networks and the bias/variance dilemma}.
\newblock \bibinfo{journal}{\emph{Neural computation}} \bibinfo{volume}{4}, \bibinfo{number}{1} (\bibinfo{year}{1992}), \bibinfo{pages}{1--58}.
\newblock


\bibitem[\protect\citeauthoryear{Geritz, Metz, Kisdi, and Mesz{\'e}na}{Geritz et~al\mbox{.}}{1997}]%
        {geritz1997dynamics}
\bibfield{author}{\bibinfo{person}{Stefan~AH Geritz}, \bibinfo{person}{Johan~AJ Metz}, \bibinfo{person}{{\'E}va Kisdi}, {and} \bibinfo{person}{G{\'e}za Mesz{\'e}na}.} \bibinfo{year}{1997}\natexlab{}.
\newblock \showarticletitle{Dynamics of adaptation and evolutionary branching}.
\newblock \bibinfo{journal}{\emph{Physical Review Letters}} \bibinfo{volume}{78}, \bibinfo{number}{10} (\bibinfo{year}{1997}).
\newblock


\bibitem[\protect\citeauthoryear{Glasmachers, Schaul, Yi, Wierstra, and Schmidhuber}{Glasmachers et~al\mbox{.}}{2010}]%
        {glasmachers2010exponential}
\bibfield{author}{\bibinfo{person}{Tobias Glasmachers}, \bibinfo{person}{Tom Schaul}, \bibinfo{person}{Sun Yi}, \bibinfo{person}{Daan Wierstra}, {and} \bibinfo{person}{J{\"u}rgen Schmidhuber}.} \bibinfo{year}{2010}\natexlab{}.
\newblock \showarticletitle{Exponential natural evolution strategies}. In \bibinfo{booktitle}{\emph{Proceedings of the 12th annual conference on Genetic and evolutionary computation}}. \bibinfo{pages}{393--400}.
\newblock


\bibitem[\protect\citeauthoryear{Gupta, Ong, and Feng}{Gupta et~al\mbox{.}}{2015}]%
        {gupta2015multifactorial}
\bibfield{author}{\bibinfo{person}{Abhishek Gupta}, \bibinfo{person}{Yew-Soon Ong}, {and} \bibinfo{person}{Liang Feng}.} \bibinfo{year}{2015}\natexlab{}.
\newblock \showarticletitle{Multifactorial evolution: toward evolutionary multitasking}.
\newblock \bibinfo{journal}{\emph{IEEE Transactions on Evolutionary Computation}} \bibinfo{volume}{20}, \bibinfo{number}{3} (\bibinfo{year}{2015}), \bibinfo{pages}{343--357}.
\newblock


\bibitem[\protect\citeauthoryear{Jakobi, Husbands, and Harvey}{Jakobi et~al\mbox{.}}{1995}]%
        {jakobi1995noise}
\bibfield{author}{\bibinfo{person}{Nick Jakobi}, \bibinfo{person}{Phil Husbands}, {and} \bibinfo{person}{Inman Harvey}.} \bibinfo{year}{1995}\natexlab{}.
\newblock \showarticletitle{Noise and the reality gap: The use of simulation in evolutionary robotics}. In \bibinfo{booktitle}{\emph{Advances in Artificial Life: Third European Conference on Artificial Life Granada, Spain, June 4--6, 1995 Proceedings 3}}. Springer, \bibinfo{pages}{704--720}.
\newblock


\bibitem[\protect\citeauthoryear{Koos, Mouret, and Doncieux}{Koos et~al\mbox{.}}{2012}]%
        {koos2012transferability}
\bibfield{author}{\bibinfo{person}{Sylvain Koos}, \bibinfo{person}{Jean-Baptiste Mouret}, {and} \bibinfo{person}{St{\'e}phane Doncieux}.} \bibinfo{year}{2012}\natexlab{}.
\newblock \showarticletitle{The transferability approach: Crossing the reality gap in evolutionary robotics}.
\newblock \bibinfo{journal}{\emph{IEEE Transactions on Evolutionary Computation}} \bibinfo{volume}{17}, \bibinfo{number}{1} (\bibinfo{year}{2012}), \bibinfo{pages}{122--145}.
\newblock


\bibitem[\protect\citeauthoryear{Liu, Yao, and Higuchi}{Liu et~al\mbox{.}}{2000}]%
        {liu2000evolutionary}
\bibfield{author}{\bibinfo{person}{Yong Liu}, \bibinfo{person}{Xin Yao}, {and} \bibinfo{person}{Tetsuya Higuchi}.} \bibinfo{year}{2000}\natexlab{}.
\newblock \showarticletitle{Evolutionary ensembles with negative correlation learning}.
\newblock \bibinfo{journal}{\emph{IEEE Transactions on Evolutionary Computation}} \bibinfo{volume}{4}, \bibinfo{number}{4} (\bibinfo{year}{2000}), \bibinfo{pages}{380--387}.
\newblock


\bibitem[\protect\citeauthoryear{Mangal, Nori, and Orso}{Mangal et~al\mbox{.}}{2019}]%
        {mangal}
\bibfield{author}{\bibinfo{person}{Ravi Mangal}, \bibinfo{person}{Aditya~V. Nori}, {and} \bibinfo{person}{Alessandro Orso}.} \bibinfo{year}{2019}\natexlab{}.
\newblock \showarticletitle{Robustness of Neural Networks: A Probabilistic and Practical Approach}. In \bibinfo{booktitle}{\emph{2019 IEEE/ACM 41st International Conference on Software Engineering: New Ideas and Emerging Results (ICSE-NIER)}}. \bibinfo{pages}{93--96}.
\newblock
\urldef\tempurl%
\url{https://doi.org/10.1109/ICSE-NIER.2019.00032}
\showDOI{\tempurl}


\bibitem[\protect\citeauthoryear{Miikkulainen}{Miikkulainen}{2010}]%
        {Miikkulainen2010}
\bibfield{author}{\bibinfo{person}{Risto Miikkulainen}.} \bibinfo{year}{2010}\natexlab{}.
\newblock \bibinfo{booktitle}{\emph{Neuroevolution}}.
\newblock \bibinfo{publisher}{Springer US}, \bibinfo{address}{Boston, MA}, \bibinfo{pages}{716--720}.
\newblock
\showISBNx{978-0-387-30164-8}
\urldef\tempurl%
\url{https://doi.org/10.1007/978-0-387-30164-8_589}
\showDOI{\tempurl}


\bibitem[\protect\citeauthoryear{Miki, Lee, Hwangbo, Wellhausen, Koltun, and Hutter}{Miki et~al\mbox{.}}{2022}]%
        {miki2022learning}
\bibfield{author}{\bibinfo{person}{Takahiro Miki}, \bibinfo{person}{Joonho Lee}, \bibinfo{person}{Jemin Hwangbo}, \bibinfo{person}{Lorenz Wellhausen}, \bibinfo{person}{Vladlen Koltun}, {and} \bibinfo{person}{Marco Hutter}.} \bibinfo{year}{2022}\natexlab{}.
\newblock \showarticletitle{Learning robust perceptive locomotion for quadrupedal robots in the wild}.
\newblock \bibinfo{journal}{\emph{Science Robotics}} \bibinfo{volume}{7}, \bibinfo{number}{62} (\bibinfo{year}{2022}), \bibinfo{pages}{eabk2822}.
\newblock


\bibitem[\protect\citeauthoryear{Milano and Nolfi}{Milano and Nolfi}{2021}]%
        {milano2021automated}
\bibfield{author}{\bibinfo{person}{Nicola Milano} {and} \bibinfo{person}{Stefano Nolfi}.} \bibinfo{year}{2021}\natexlab{}.
\newblock \showarticletitle{Automated curriculum learning for embodied agents a neuroevolutionary approach}.
\newblock \bibinfo{journal}{\emph{Scientific reports}} \bibinfo{volume}{11}, \bibinfo{number}{1} (\bibinfo{year}{2021}), \bibinfo{pages}{8985}.
\newblock


\bibitem[\protect\citeauthoryear{Morse and Stanley}{Morse and Stanley}{2016}]%
        {morse2016simple}
\bibfield{author}{\bibinfo{person}{Gregory Morse} {and} \bibinfo{person}{Kenneth~O Stanley}.} \bibinfo{year}{2016}\natexlab{}.
\newblock \showarticletitle{Simple evolutionary optimization can rival stochastic gradient descent in neural networks}. In \bibinfo{booktitle}{\emph{Proceedings of the Genetic and Evolutionary Computation Conference 2016}}. \bibinfo{pages}{477--484}.
\newblock


\bibitem[\protect\citeauthoryear{Narvekar, Peng, Leonetti, Sinapov, Taylor, and Stone}{Narvekar et~al\mbox{.}}{2020}]%
        {narvekar2020curriculum}
\bibfield{author}{\bibinfo{person}{Sanmit Narvekar}, \bibinfo{person}{Bei Peng}, \bibinfo{person}{Matteo Leonetti}, \bibinfo{person}{Jivko Sinapov}, \bibinfo{person}{Matthew~E Taylor}, {and} \bibinfo{person}{Peter Stone}.} \bibinfo{year}{2020}\natexlab{}.
\newblock \showarticletitle{Curriculum learning for reinforcement learning domains: A framework and survey}.
\newblock \bibinfo{journal}{\emph{The Journal of Machine Learning Research}} \bibinfo{volume}{21}, \bibinfo{number}{1} (\bibinfo{year}{2020}), \bibinfo{pages}{7382--7431}.
\newblock


\bibitem[\protect\citeauthoryear{Nygaard, Martin, Torresen, Glette, and Howard}{Nygaard et~al\mbox{.}}{2021}]%
        {nygaard2021real}
\bibfield{author}{\bibinfo{person}{T{\o}nnes~F Nygaard}, \bibinfo{person}{Charles~P Martin}, \bibinfo{person}{Jim Torresen}, \bibinfo{person}{Kyrre Glette}, {and} \bibinfo{person}{David Howard}.} \bibinfo{year}{2021}\natexlab{}.
\newblock \showarticletitle{Real-world embodied AI through a morphologically adaptive quadruped robot}.
\newblock \bibinfo{journal}{\emph{Nature Machine Intelligence}} \bibinfo{volume}{3}, \bibinfo{number}{5} (\bibinfo{year}{2021}), \bibinfo{pages}{410--419}.
\newblock


\bibitem[\protect\citeauthoryear{Pagliuca, Milano, and Nolfi}{Pagliuca et~al\mbox{.}}{2020}]%
        {pagliuca2020efficacy}
\bibfield{author}{\bibinfo{person}{Paolo Pagliuca}, \bibinfo{person}{Nicola Milano}, {and} \bibinfo{person}{Stefano Nolfi}.} \bibinfo{year}{2020}\natexlab{}.
\newblock \showarticletitle{Efficacy of modern neuro-evolutionary strategies for continuous control optimization}.
\newblock \bibinfo{journal}{\emph{Frontiers in Robotics and AI}}  \bibinfo{volume}{7} (\bibinfo{year}{2020}), \bibinfo{pages}{98}.
\newblock


\bibitem[\protect\citeauthoryear{Pfeifer and Bongard}{Pfeifer and Bongard}{2006}]%
        {pfeifer2006body}
\bibfield{author}{\bibinfo{person}{Rolf Pfeifer} {and} \bibinfo{person}{Josh Bongard}.} \bibinfo{year}{2006}\natexlab{}.
\newblock \bibinfo{booktitle}{\emph{How the body shapes the way we think: a new view of intelligence}}.
\newblock \bibinfo{publisher}{MIT press}.
\newblock


\bibitem[\protect\citeauthoryear{Putter and Nitschke}{Putter and Nitschke}{2017}]%
        {putter2017evolving}
\bibfield{author}{\bibinfo{person}{Ruben Putter} {and} \bibinfo{person}{Geoff Nitschke}.} \bibinfo{year}{2017}\natexlab{}.
\newblock \showarticletitle{Evolving morphological robustness for collective robotics}. In \bibinfo{booktitle}{\emph{2017 IEEE Symposium Series on Computational Intelligence (SSCI)}}. IEEE, \bibinfo{pages}{1--8}.
\newblock


\bibitem[\protect\citeauthoryear{Putter and Nitschke}{Putter and Nitschke}{2018}]%
        {8628627}
\bibfield{author}{\bibinfo{person}{Ruben Putter} {and} \bibinfo{person}{Geoff Nitschke}.} \bibinfo{year}{2018}\natexlab{}.
\newblock \showarticletitle{Objective versus Non-Objective Search in Evolving Morphologically Robust Robot Controllers}. In \bibinfo{booktitle}{\emph{2018 IEEE Symposium Series on Computational Intelligence (SSCI)}}. \bibinfo{pages}{2033--2040}.
\newblock
\urldef\tempurl%
\url{https://doi.org/10.1109/SSCI.2018.8628627}
\showDOI{\tempurl}


\bibitem[\protect\citeauthoryear{Rajeswaran, Lowrey, Todorov, and Kakade}{Rajeswaran et~al\mbox{.}}{2017}]%
        {rajeswaran2017towards}
\bibfield{author}{\bibinfo{person}{Aravind Rajeswaran}, \bibinfo{person}{Kendall Lowrey}, \bibinfo{person}{Emanuel~V Todorov}, {and} \bibinfo{person}{Sham~M Kakade}.} \bibinfo{year}{2017}\natexlab{}.
\newblock \showarticletitle{Towards generalization and simplicity in continuous control}.
\newblock \bibinfo{journal}{\emph{Advances in Neural Information Processing Systems}}  \bibinfo{volume}{30} (\bibinfo{year}{2017}).
\newblock


\bibitem[\protect\citeauthoryear{Raviv, Lupyan, and Green}{Raviv et~al\mbox{.}}{2022}]%
        {raviv2022variability}
\bibfield{author}{\bibinfo{person}{Limor Raviv}, \bibinfo{person}{Gary Lupyan}, {and} \bibinfo{person}{Shawn~C Green}.} \bibinfo{year}{2022}\natexlab{}.
\newblock \showarticletitle{How variability shapes learning and generalization}.
\newblock \bibinfo{journal}{\emph{Trends in cognitive sciences}} \bibinfo{volume}{26}, \bibinfo{number}{6} (\bibinfo{year}{2022}), \bibinfo{pages}{462--483}.
\newblock


\bibitem[\protect\citeauthoryear{Risi and Stanley}{Risi and Stanley}{2013}]%
        {risi2013confronting}
\bibfield{author}{\bibinfo{person}{Sebastian Risi} {and} \bibinfo{person}{Kenneth~O Stanley}.} \bibinfo{year}{2013}\natexlab{}.
\newblock \showarticletitle{Confronting the challenge of learning a flexible neural controller for a diversity of morphologies}. In \bibinfo{booktitle}{\emph{Proceedings of the 15th annual conference on Genetic and evolutionary computation}}. \bibinfo{pages}{255--262}.
\newblock


\bibitem[\protect\citeauthoryear{Salimans, Ho, Chen, Sidor, and Sutskever}{Salimans et~al\mbox{.}}{2017}]%
        {salimans2017evolution}
\bibfield{author}{\bibinfo{person}{Tim Salimans}, \bibinfo{person}{Jonathan Ho}, \bibinfo{person}{Xi Chen}, \bibinfo{person}{Szymon Sidor}, {and} \bibinfo{person}{Ilya Sutskever}.} \bibinfo{year}{2017}\natexlab{}.
\newblock \showarticletitle{Evolution strategies as a scalable alternative to reinforcement learning}.
\newblock \bibinfo{journal}{\emph{arXiv preprint arXiv:1703.03864}} (\bibinfo{year}{2017}).
\newblock


\bibitem[\protect\citeauthoryear{Schaul}{Schaul}{2012}]%
        {schaul2012benchmarking}
\bibfield{author}{\bibinfo{person}{Tom Schaul}.} \bibinfo{year}{2012}\natexlab{}.
\newblock \showarticletitle{Benchmarking exponential natural evolution strategies on the noiseless and noisy black-box optimization testbeds}. In \bibinfo{booktitle}{\emph{Proceedings of the 14th annual conference companion on Genetic and evolutionary computation}}. \bibinfo{pages}{213--220}.
\newblock


\bibitem[\protect\citeauthoryear{Stanley, Clune, Lehman, and Miikkulainen}{Stanley et~al\mbox{.}}{2019}]%
        {stanley2019designing}
\bibfield{author}{\bibinfo{person}{Kenneth~O Stanley}, \bibinfo{person}{Jeff Clune}, \bibinfo{person}{Joel Lehman}, {and} \bibinfo{person}{Risto Miikkulainen}.} \bibinfo{year}{2019}\natexlab{}.
\newblock \showarticletitle{Designing neural networks through neuroevolution}.
\newblock \bibinfo{journal}{\emph{Nature Machine Intelligence}} \bibinfo{volume}{1}, \bibinfo{number}{1} (\bibinfo{year}{2019}), \bibinfo{pages}{24--35}.
\newblock


\bibitem[\protect\citeauthoryear{Such, Madhavan, Conti, Lehman, Stanley, and Clune}{Such et~al\mbox{.}}{2017}]%
        {such2017deep}
\bibfield{author}{\bibinfo{person}{Felipe~Petroski Such}, \bibinfo{person}{Vashisht Madhavan}, \bibinfo{person}{Edoardo Conti}, \bibinfo{person}{Joel Lehman}, \bibinfo{person}{Kenneth~O Stanley}, {and} \bibinfo{person}{Jeff Clune}.} \bibinfo{year}{2017}\natexlab{}.
\newblock \showarticletitle{Deep neuroevolution: Genetic algorithms are a competitive alternative for training deep neural networks for reinforcement learning}.
\newblock \bibinfo{journal}{\emph{arXiv preprint arXiv:1712.06567}} (\bibinfo{year}{2017}).
\newblock


\bibitem[\protect\citeauthoryear{Telikani, Tahmassebi, Banzhaf, and Gandomi}{Telikani et~al\mbox{.}}{2021}]%
        {telikani2021evolutionary}
\bibfield{author}{\bibinfo{person}{Akbar Telikani}, \bibinfo{person}{Amirhessam Tahmassebi}, \bibinfo{person}{Wolfgang Banzhaf}, {and} \bibinfo{person}{Amir~H Gandomi}.} \bibinfo{year}{2021}\natexlab{}.
\newblock \showarticletitle{Evolutionary machine learning: A survey}.
\newblock \bibinfo{journal}{\emph{ACM Computing Surveys (CSUR)}} \bibinfo{volume}{54}, \bibinfo{number}{8} (\bibinfo{year}{2021}), \bibinfo{pages}{1--35}.
\newblock


\bibitem[\protect\citeauthoryear{Toklu, Atkinson, Micka, Liskowski, and Srivastava}{Toklu et~al\mbox{.}}{2023}]%
        {evotorch2023arxiv}
\bibfield{author}{\bibinfo{person}{Nihat~Engin Toklu}, \bibinfo{person}{Timothy Atkinson}, \bibinfo{person}{Vojt\v{e}ch Micka}, \bibinfo{person}{Pawe\l{} Liskowski}, {and} \bibinfo{person}{Rupesh~Kumar Srivastava}.} \bibinfo{year}{2023}\natexlab{}.
\newblock \showarticletitle{{EvoTorch}: Scalable Evolutionary Computation in {Python}}.
\newblock \bibinfo{journal}{\emph{arXiv preprint}} (\bibinfo{year}{2023}).
\newblock
\newblock
\shownote{https://arxiv.org/abs/2302.12600}.


\bibitem[\protect\citeauthoryear{Valentim, Louren{\c{c}}o, and Antunes}{Valentim et~al\mbox{.}}{2022}]%
        {valentim2022adversarial}
\bibfield{author}{\bibinfo{person}{In{\^e}s Valentim}, \bibinfo{person}{Nuno Louren{\c{c}}o}, {and} \bibinfo{person}{Nuno Antunes}.} \bibinfo{year}{2022}\natexlab{}.
\newblock \showarticletitle{Adversarial Robustness Assessment of NeuroEvolution Approaches}. In \bibinfo{booktitle}{\emph{2022 IEEE Congress on Evolutionary Computation (CEC)}}. IEEE, \bibinfo{pages}{1--8}.
\newblock


\bibitem[\protect\citeauthoryear{Wang, Lehman, Clune, and Stanley}{Wang et~al\mbox{.}}{2019}]%
        {wang2019poet}
\bibfield{author}{\bibinfo{person}{Rui Wang}, \bibinfo{person}{Joel Lehman}, \bibinfo{person}{Jeff Clune}, {and} \bibinfo{person}{Kenneth~O Stanley}.} \bibinfo{year}{2019}\natexlab{}.
\newblock \showarticletitle{Poet: open-ended coevolution of environments and their optimized solutions}. In \bibinfo{booktitle}{\emph{Proceedings of the Genetic and Evolutionary Computation Conference}}. \bibinfo{pages}{142--151}.
\newblock


\bibitem[\protect\citeauthoryear{Wang, Wu, Fu, Fu, Zhang, Chang, and Wang}{Wang et~al\mbox{.}}{2022}]%
        {wang2022curriculum}
\bibfield{author}{\bibinfo{person}{Yuxing Wang}, \bibinfo{person}{Shuang Wu}, \bibinfo{person}{Haobo Fu}, \bibinfo{person}{Qiang Fu}, \bibinfo{person}{Tiantian Zhang}, \bibinfo{person}{Yongzhe Chang}, {and} \bibinfo{person}{Xueqian Wang}.} \bibinfo{year}{2022}\natexlab{}.
\newblock \showarticletitle{Curriculum-based co-design of morphology and control of voxel-based soft robots}. In \bibinfo{booktitle}{\emph{The Eleventh International Conference on Learning Representations}}.
\newblock


\bibitem[\protect\citeauthoryear{Wierstra, Schaul, Glasmachers, Sun, Peters, and Schmidhuber}{Wierstra et~al\mbox{.}}{2014}]%
        {wierstra2014natural}
\bibfield{author}{\bibinfo{person}{Daan Wierstra}, \bibinfo{person}{Tom Schaul}, \bibinfo{person}{Tobias Glasmachers}, \bibinfo{person}{Yi Sun}, \bibinfo{person}{Jan Peters}, {and} \bibinfo{person}{J{\"u}rgen Schmidhuber}.} \bibinfo{year}{2014}\natexlab{}.
\newblock \showarticletitle{Natural evolution strategies}.
\newblock \bibinfo{journal}{\emph{The Journal of Machine Learning Research}} \bibinfo{volume}{15}, \bibinfo{number}{1} (\bibinfo{year}{2014}), \bibinfo{pages}{949--980}.
\newblock


\bibitem[\protect\citeauthoryear{Yaman, Iacca, Mocanu, Coler, Fletcher, and Pechenizkiy}{Yaman et~al\mbox{.}}{2021}]%
        {yaman2021evolving}
\bibfield{author}{\bibinfo{person}{Anil Yaman}, \bibinfo{person}{Giovanni Iacca}, \bibinfo{person}{Decebal~Constantin Mocanu}, \bibinfo{person}{Matt Coler}, \bibinfo{person}{George Fletcher}, {and} \bibinfo{person}{Mykola Pechenizkiy}.} \bibinfo{year}{2021}\natexlab{}.
\newblock \showarticletitle{Evolving plasticity for autonomous learning under changing environmental conditions}.
\newblock \bibinfo{journal}{\emph{Evolutionary computation}} \bibinfo{volume}{29}, \bibinfo{number}{3} (\bibinfo{year}{2021}), \bibinfo{pages}{391--414}.
\newblock


\bibitem[\protect\citeauthoryear{Yaman, Mocanu, Iacca, Fletcher, and Pechenizkiy}{Yaman et~al\mbox{.}}{2018}]%
        {yaman2018limited}
\bibfield{author}{\bibinfo{person}{Anil Yaman}, \bibinfo{person}{Decebal~Constantin Mocanu}, \bibinfo{person}{Giovanni Iacca}, \bibinfo{person}{George Fletcher}, {and} \bibinfo{person}{Mykola Pechenizkiy}.} \bibinfo{year}{2018}\natexlab{}.
\newblock \showarticletitle{Limited evaluation cooperative co-evolutionary differential evolution for large-scale neuroevolution}. In \bibinfo{booktitle}{\emph{Proceedings of the Genetic and Evolutionary Computation Conference}}. \bibinfo{pages}{569--576}.
\newblock


\bibitem[\protect\citeauthoryear{Yang, Yu, You, Steinhardt, and Ma}{Yang et~al\mbox{.}}{2020}]%
        {yang2020rethinking}
\bibfield{author}{\bibinfo{person}{Zitong Yang}, \bibinfo{person}{Yaodong Yu}, \bibinfo{person}{Chong You}, \bibinfo{person}{Jacob Steinhardt}, {and} \bibinfo{person}{Yi Ma}.} \bibinfo{year}{2020}\natexlab{}.
\newblock \showarticletitle{Rethinking bias-variance trade-off for generalization of neural networks}. In \bibinfo{booktitle}{\emph{International Conference on Machine Learning}}. PMLR, \bibinfo{pages}{10767--10777}.
\newblock


\end{thebibliography}
